# Rearranging the Familiar: Testing Compositional Generalization in Recurrent Networks

**João Loula**
École Polytechnique
Facebook AI Research
joaoloula@fb.com

**Marco Baroni**
Facebook AI Research
mbaroni@fb.com

**Brenden M. Lake**
New York University
Facebook AI Research
brenden@nyu.edu

## Abstract

Systematic compositionality is the ability to recombine meaningful units with regular and predictable outcomes, and it's seen as key to the human capacity for generalization in language. Recent work (Lake and Baroni, 2018) has studied systematic compositionality in modern seq2seq models using generalization to novel navigation instructions in a grounded environment as a probing tool. Lake and Baroni's main experiment required the models to quickly bootstrap the meaning of new words. We extend this framework here to settings where the model needs only to recombine well-trained functional words (such as "*around*" and "*right*") in novel contexts. Our findings confirm and strengthen the earlier ones: seq2seq models can be impressively good at generalizing to novel combinations of previously-seen input, but only when they receive extensive training on the specific pattern to be generalized (e.g., generalizing from many examples of "X *around right*" to "*jump around right*"), while failing when generalization requires novel application of compositional rules (e.g., inferring the meaning of "*around right*" from those of "*right*" and "*around*").

## 1 Introduction

Human language learning enjoys a good kind of combinatorial explosion — if a person knows the meaning of "*to run*" and that of "*slowly*", she can immediately understand what it means "*to run slowly*", even if she has never uttered or heard this expression before. This is an example of *compositionality*, the algebraic capacity to understand and produce novel combinations from known components (Montague, 1970). This principle helps to explain how, when acquiring a language, we can quickly bootstrap to a potentially infinite number of expressions from very limited training data (Chomsky, 1957).

Neural networks have recently been successfully applied to many tasks requiring considerable generalization abilities (LeCun et al., 2015), including applications in the domain of natural language (Goldberg, 2017). However, it has also been observed that they require a very large number of training examples to succeed, which suggests that they lack compositional abilities (Lake et al., 2017). There has been a substantial earlier debate on the extent to which neural networks display some degree of compositional generalization (e.g., Fodor and Pylyshyn, 1988; Christiansen and Chater, 1994; Marcus, 1998; Phillips, 1998; Chang, 2002; Marcus, 2003; van der Velde et al., 2004; Bowers et al., 2009; Botvinick and Plaut, 2009; Brakel and Frank, 2009; Frank, 2014). Recently, Lake and Baroni (2018) revisited these issues in light of the latest advances in deep neural networks for natural language processing.

The authors introduced the SCAN dataset for studying compositionality in sequence-to-sequence (seq2seq) neural network models (Sutskever et al., 2014). SCAN is a simple language-driven navigation environment that supports one-shot learning experiments, where the trained agent must execute test commands that it has never encountered in training, but are assembled from the same components as the training commands.

Lake and Baroni found that state-of-the-art recurrent neural networks (RNNs) showed impressive zero-shot generalization capabilities when commands were arbitrarily split between train and test set, but they failed in cases that required *systematic* compositionality, that is, to extract algebraic composition rules from the training examples. To begin with, RNNs failed when they had to generalize to commands requiring longer action sequences to be executed. This is not too surprising, as longer sequences are notoriously challeng-ing for seq2seq models (Cho et al., 2014). More interestingly, Lake and Baroni found that RNNs do not correctly generalize the usage of a new action verb (shown in isolation during training) to contexts that are familiar from other verbs. In other words, RNNs fail the following basic compositionality test: Even after they acquired the meaning of "*to run again*" and "*to dax*", they do not understand "*to dax again*" on first encounter.

As Lake and Baroni show, the generalization problem is linked to the fact that RNNs fail to learn a representation (an embedding) for the new verb ("*to dax*") that is similar to those of known verbs ("*to run*", "*to look*"), and consequently it cannot rely on similarity information to correctly generalize verb usage. This is arguably more of an instance of the problem of quickly learning meaningful new-word embeddings (Herbelot and Baroni, 2017; Lampinen and McClelland, 2017), than strictly a failure of compositionality.

In this paper, we repurpose SCAN to test another kind of compositionality, namely one that requires combining highly familiar words in new ways to create novel meaning. As illustrated above, this is what we do when we combine a functional term such as "*slowly*" with the verb "*to run*" to obtain the phrase "*to run slowly*". Or, in terms of the SCAN commands that we test here, this is what is required to understand an expression such as "*jump around right*" when the meanings of "*jump*", "*right*" and "*around*" are known.

Our results confirm and strengthen the conclusions of Lake and Baroni. On the one hand, RNNs do show a considerable degree of generalization in our experiments as well. However, their performance dramatically decreases as the difference between training and testing becomes more systematic, even though all test examples could be correctly processed by relying on simple composition rules amply illustrated in the training data.

## 2 Generalizing functional terms with SCAN

The SCAN dataset (Lake and Baroni, 2018) presents the problem of translating commands from a simplified natural language to a sequence of actions, framed as a seq2seq task (Sutskever et al., 2014). The commands are generated by a phrase-structure grammar and then converted into actions by a semantic interpretation function.

By way of illustration, let us take a prototypical SCAN command like "*turn right twice and jump around left*". This command's building blocks are "*turn right*" and "Primitive *around left*", which are part of SCAN's 12 *templates*, a collection of base expressions that present a great deal of symmetry over actions, spatial terms, and manner adverbs (Table 1). Some of these templates can operate over different Primitives ("*jump*", "*walk*", "*run*", or "*look*"), mapping them systematically to their correspondent output [Primitive] ("JUMP", "WALK", "RUN", and "LOOK"). The templates can in turn be combined using the conjunctions "*and*" and "*after*" and quantified by "*twice*" and "*thrice*" for a total of 20,910 commands: these include things like "*walk left after look opposite left*", "*turn around right thrice*", "*jump right and run left*" etc.

Lake and Baroni (2018) present three experiments based on different train-test splits: a random split, a split where the test set contains commands requiring longer action sequences than the training set, and a split where the test set contains commands with compositions of primitives of which few examples exist in the training set (in the limit, the primitives are only presented in isolation). Their main conclusion is that neural networks, though surprisingly good at zero-shot generalization to novel commands, are still far from systematic compositionality. In the first split, networks are able to achieve high accuracy with relatively few training examples. In the second and third ones, where the training/testing gap is larger yet there exist systematic rules linking the training and test sets, the same models fail.

The main contribution of this paper is showing how the SCAN dataset can be repurposed to analyze compositionality with known functional terms used in new contexts, where it is not a matter of quickly learning a new embedding as in the original primitive generalization experiment, but rather of adequately recombining familiar words. The key insight is that manner adverbs in the dataset, such as "*around*" and "*opposite*", act as second-order modifiers, operating over the spatial modifiers "*left*" and "*right*" and the primitives "*look*", "*walk*", "*run*" and "*jump*". This opens up the possibility of splitting the dataset such that examples like "*walk left*", "*walk right*", and "*jump around left*" are seen in the training set, and at test time the network must piece these together to interpret commands like "*jump around right*", which contain only extensively seen words, but presented in a new context. In other words, the network must internalize the symmetry between the terms "*left*" and "*right*" that is evident across SCAN (by comparing templates 1 and 2, templates 3 and 4, etc. in Table 1), and use it to learn abstract rules for higher-order modifiers such as *"opposite"*.

| Template | Command | Target |
|---|---|---|
| 1 | "*turn left*" | LTURN |
| 2 | "*turn right*" | RTURN |
| 3 | "*Primitive left*" | LTURN [Primitive] |
| 4 | "*Primitive right*" | RTURN [Primitive] |
| 5 | "*turn opposite left*" | LTURN LTURN |
| 6 | "*turn opposite right*" | RTURN RTURN |
| 7 | "*Primitive opposite left*" | LTURN LTURN [Primitive] |
| 8 | "*Primitive opposite right*" | RTURN RTURN [Primitive] |
| 9 | "*turn around left*" | LTURN LTURN LTURN LTURN |
| 10 | "*turn around right*" | RTURN RTURN RTURN RTURN |
| 11 | "*Primitive around left*" | LTURN [Primitive] LTURN [Primitive] LTURN [Primitive] LTURN [Primitive] |
| 12 | "*Primitive around right*" | RTURN [Primitive] RTURN [Primitive] RTURN [Primitive] RTURN [Primitive] |

Table 1: All command templates in the SCAN dataset, along with the target output. Here, "Primitive" can stand for "*jump*", "*walk*", "*run*", or "*look*", with the corresponding output [Primitive] being "JUMP", "WALK", "RUN", or "LOOK".

## 3 Experiments

All experiments were run using the overall best neural network from Lake and Baroni (2018): a seq2seq 2-layer, 200-unit LSTM with 50% dropout (Figure 1). The values of all other hyperparameters were those specified by Lake and Baroni. This model was very successful in their basic, random-split experiment, where it achieved 99.8% accuracy. We also tried the best attention-augmented model from Lake and Baroni, but it was outperformed by the overall-best in all experiments, and is thus omitted here. All test-set accuracies are reported with mean and standard deviation across 5 runs: in experiments where the splits were created by random sampling, each run corresponds to a different sample. Though the size of the training set varies across conditions, the training regime is always fixed at 100k presentations (approximately 5 epochs for the condition with the largest training set): in practice, this was sufficient for near-perfect training set accuracy in all conditions[1].

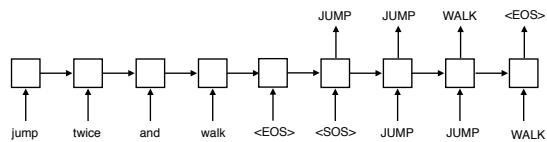

Figure 1: Illustration of the sequence-to-sequence model operating on the SCAN dataset. The network takes a command such as "*jump twice and walk*" and must convert it to a sequence of actions, in this case "JUMP", "JUMP", "WALK". Reproduced from Lake and Baroni (2018) with permission.

**Experiment 1: Generalizing to novel templates**

In order to probe the network's ability to recombine well-trained words as well as to assess the factors that render that task easier or harder, we compared performance across 4 different train-test splits. In the first one we leave out examples containing the subcommand "*jump around right*" (a specific instance of Template 12) whereas in the other 3 we leave out all instances of different templates described in-depth in Table 1. In all of the splits, the network is tasked with generalizing to novel commands involving "*right*" by exploiting the "*left*"/"*right*" symmetry in the training set and/or the distributional similarity among primitives. We present the splits in order of conjectured increasing complexity, in terms of systematic gaps between training and test sets. Table 2 shows examples of commands in the training and test set for the different conditions.

---
[1] All train-test splits are available along with the original SCAN dataset at: https://github.com/brendenlake/SCAN.

- **jump around right**: The test set consists of all commands containing the phrase "*jump around right*", while all remaining commands are in the training set, including uses of "*jump around left*" and "Primitive *around right*" for the other primitives. The network is thus exposed to plenty of evidence that *"jump"* has the same distribution as the other primitives (thus, it should easily discover the similarity of "*jump*" to the other primitives), and it sees many instances of the "Primitive *around right*" template with all other primitive fillers but *"jump"*.

- **Primitive *right***: The test set consists of all commands containing "Primitive *right*" (Template 4 in Table 1), with all remaining templates (and their conjunctions and quantifications) in the training set. In this case, the network is exposed to "Primitive *left*" and many examples illustrating the *"left"/"right"* symmetry during training, and it must bootstrap to the *simplest* usage of *"right"* at test time.

- **Primitive *opposite right***: The test set consists of all commands containing templates of the form "Primitive *opposite right*" (Template 8), with the remaining templates (and their conjunctions and quantifications) in the training set. Here, the network is never exposed to the "Primitive *opposite right*" template with *any* primitive filler, and it has to bootstrap the combined effect of *"opposite"* and *"right"* based on seeing them applied independently, plus the *"left"/"right"* symmetry.

- **Primitive *around right***: The test set consists of all commands containing templates of the form "Primitive *around right*" (Template 12), with the remaining templates in the training set. This is analogous to "Primitive *opposite right*", but requires executing a longer action sequence due to the different SCAN semantics of *"opposite"* (two turning+Primitive steps to turn in the opposite direction) vs. *"around"* (four turning+Primitive steps to perform a full roundabout, refer to Table 1).

Observe that "*turn*" in SCAN has a different semantics from the other actions verbs (see Table 1). We found that removing all commands where "*turn*" appeared in the target expression (e.g. "*turn around right*" in the "Primitive *around right*" condition, "*turn opposite right*" in the "Primitive *opposite right*" condition etc.) from both training and test sets systematically increased accuracy, and we thus report results in this setup.

**Results:** A summary of the results is presented in Table 3. We see that the network had no problem generalizing to *"jump around right"* when being exposed to all commands containing this template with all other possible fillers. This confirms Lake and Baroni's result that modern RNNs do to some extent generalize to new constructions. However, the remaining results also confirm their finding of a lack of systematicity in generalization.

Interestingly, the poor performance in the "Primitive *right*" condition shows that generalization is problematic for RNNs not only when they have to bootstrap to *longer* constructions, but also when they have to systematically generalize to *shorter* ones (a network exposed to "*run left*", *"run opposite right"*, *"jump left"*, *"jump around right"* etc. fails to execute *"run right"* or *"jump right"*).

The dramatic difference in accuracy between training without "*around right*" commands vs. training on all templates except "*jump around right*" commands (2.46% vs. 98.43%) points to the network being able to generalize the application of "*around right*" across primitives, but not being able to directly apply "*right*" and "*around*" to a primitive, without having seen them presented together. The failure modes here further showcase, qualitatively, the lack of systematicity. For instance, though the network correctly interprets the complex expression "*jump right after walk around right*", it fails to do so for the subcommand "*walk around right*", where it flips one of the four "*right*" turns for a "*left*" one.

Surprisingly, the network, while still far from perfect, has considerably higher accuracy (47.62%) when generalizing to "*opposite right*," a simpler command of the same nature as "*around right*." This suggests that memory factors (learning to repeat the relevant steps 4 times instead of 2) interact with the network ability to extract the right patterns.

| Condition | Example train commands | Example test commands |
|---|---|---|
| **jump around right** | "jump left", "jump around left", "walk around right" | "jump around right", "jump around right and walk" |
| **Primitive** *right* | "jump left", "walk around right" | "jump right", "walk right" |
| **Primitive** *opposite right* | "jump left", "jump opposite left", "walk right" | "jump opposite right", "walk opposite right" |
| **Primitive** *around right* | "jump left", "jump around left", "walk right" | "jump around right", "walk around right" |

Table 2: Example train and test commands for different conditions of Experiment 1. Note that the train commands are meant to illustrate relevant constructions, but the training set always contains all possible commands not in the test set.

| Condition | Acc. ± s.d. | Test size |
|---|---|---|
| **jump around right** | 98.43% ±0.54% | 1,173 |
| **Primitive** *right* | 23.49% ±8.09% | 4,476 |
| **Primitive** *opposite right* | 47.62% ±17.72% | 4,476 |
| **Primitive** *around right* | 2.46% ±2.68% | 4,476 |

Table 3: Experiment 1: test set accuracy mean and standard deviation for different train-test splits on the SCAN dataset. Test set sizes are also reported.

### Experiment 2: Impact of filler variety in learning a complex template

One interesting result of Experiment 1 is that the number of distinct primitive fillers of a template that the network sees in training affects its ability to generalize the template, as shown by the striking performance difference between the "Primitive *around right*" (0 fillers of the relevant template seen in training, very low accuracy) and "*jump around right*" conditions (3 fillers seen in training, near-perfect generalization). In Experiment 2, we take a detailed look at this phenomenon by varying the number of primitive fillers for this template (Template 12 in Table 1) that the model observes during training, with the goal of learning the full abstract template. We fix the test set across all conditions by making it consist only of the commands containing the expression "*jump around right*" — this allows for more direct comparison across the conditions. Again, commands containing the expression "*turn around right*" were removed to avoid interference. The different conditions for this experiment are, in order of decreasing difficulty:

- *0 fillers*: The training set contains **no** examples of Template 12, e.g., no command of the form "Primitive *around right*". It does contain all other complete templates (1-11) in Table 1.

- *1 filler*: The training set has commands containing "*look around right*" for Template 12 as well as all other complete templates in Table 1.[2]

- *2 fillers*: The training set has commands containing "*look around right*" and "*walk around right*" for Template 12 as well as all other complete templates in Table 1.

- *3 fillers*: The training set has commands containing the templates "*look around right*", "*walk around right*" and "*run around right*" for Template 12 as well as all other complete templates in Table 1.

Each new template corresponds to roughly an additional 1,100 distinct examples in the training set.[3] Note that the actual primitives chosen for each condition do not matter, as their distribution is identical.

**Results:** A summary of the results is shown in Figure 2. We observe that the network only needs examples of 1 primitive filler to start generalizing almost perfectly to other fillers of the template. So, crucially, the network seems able to perform some analogical generalization from a verb to the other in the *"around right"* context, but not to productively apply the *"right"* and *"around"* rules to a verb, when their combined effect has never been observed.

---

[2] If a command contains both "look around right" and "Primitive + *around right*" for another primitive, then that command is held out. This is true of the other conditions as well.

[3] We remind the reader that, though the number of distinct examples in the training set varies across conditions, the total number of presentations seen during the training regime is fixed at 100k.

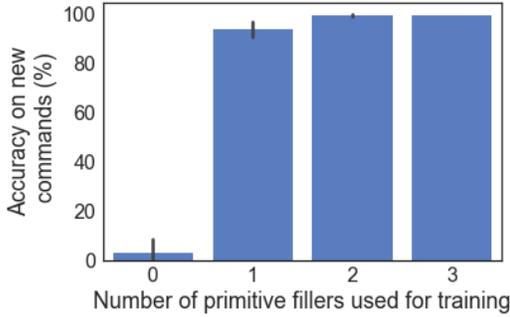

Figure 2: Experiment 2: accuracy on held-out commands containing "*jump around right*" after training on sets including a different number of commands of the form "Primitive *around right*". Error bars are bootstrapped 95% confidence intervals.

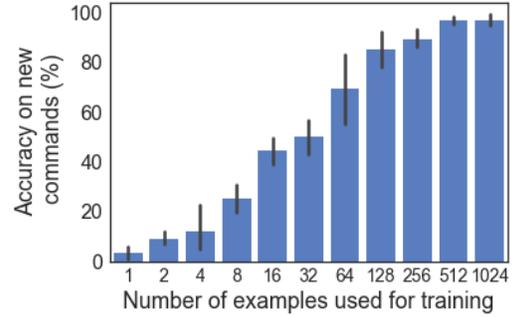

Figure 3: Experiment 3: accuracy on held-out commands containing "*jump around right*" after training on sets including a different number of commands containing "*walk around right*". Error bars are bootstrapped 95% confidence intervals.

### Experiment 3: Impact of number of distinct training examples in learning a complex template

We consider here a further level of granularity. Adding one additional primitive filler, as we did in Experiment 2, corresponds to about 1,100 additional distinct training examples. Are they all needed, or is it sufficient to observe the target complex template in a smaller number of examples? This question is the subject of Experiment 3. In order to analyze the sample complexity of the model's generalizations, we now take the *0 filler* condition from Experiment 2 and progressively add examples from the *1 filler* condition. More precisely, we randomly add 1, 2, 4, 8, 16, 32, 64, 128, 256, 512, or 1,024 commands containing "*look around right*", but no other command of the form "Primitive *around right*", to the training set of the *0 filler* condition. As before, all other templates (1-11) and their conjunctions and quantifications are also provided during training. Note that 1,024 is approximately the difference in distinct examples between the *0* and *1 filler* conditions, such that this experiment spans the entire range from one to the other.

**Results:** A summary of the results is shown in Figure 3. On the one hand, the sample complexity with which performance ramps up is quite impressive, being at a respectable 70% with 64 additional examples and peaking at 512 examples. On the other hand, the very fact that performance increases gradually, and that it takes so long for the network to peak points to a failure to generalize systematically: instead of piecing together the general rule, the network seems to be rather accumulating evidence for some specific cases.

## 4 Conclusion

Our findings complement those of Lake and Baroni (2018) now in a setting where, instead of having to learn a new embedding, the network needs only to recombine well-trained functional words, such as "*right*" and "*around*". The results show the impressive generalization capabilities of seq2seq models, correctly interpreting complex new combinations of previously seen commands, but also their lack of systematicity. On the one hand, as shown in Experiment 2, the fact that the network is able to correctly generalize to new constructions of the form "Primitive *around right*" after only seeing this template with one filler primitive is quite impressive. On the other hand, Experiment 1 suggests that this generalization is not based on the network being able to combine systematic composition rules associated to the functional terms "*right*" and "*around*". Experiment 3 further confirms that generalization is not systematic in nature, and that the network still needs to be shown a wealth of additional examples in the same context as the test set in order to achieve it, even though it has already observed ample evidence for all the test words in the training set.

Future directions include probing what kind of training set evidence is crucial for systematic generalization, and how the ability to generalize in this manner differs across different kinds of commands (primitives, manner adverbs, spatial expressions, etc.). Further empirical investigations might focus on generalization of functional terms in real-life seq2seq tasks, such as machine translation. On the modeling side, we need to study what are the right priors to encode in seq2seq models to endow them with the ability of systematic generalization without losing their generality.